\documentclass[10pt,twocolumn,letterpaper]{article}

\usepackage{iccv}
\usepackage{times}
\usepackage{epsfig}
\usepackage{graphicx}
\usepackage{amsmath}
\usepackage{amssymb}
\usepackage{enumitem}
\usepackage{caption}
\usepackage{subcaption}

\usepackage[breaklinks=true,bookmarks=false]{hyperref}

\iccvfinalcopy 

\def\iccvPaperID{1234} 

\begin{document}

\title{Dual Path Multi-Scale Fusion Networks with Attention for Crowd Counting}

\author{Liang Zhu, Zhijian Zhao, Chao Lu, Yining Lin, Yao Peng, Tangren Yao\\
Qiniu AtLab\\
{\tt\small \{zhuliang, zhaozhijian, luchao, linyining, pengyao, yaotangren\}@qiniu.com}
}

\maketitle

\begin{abstract}
The task of crowd counting in varying density scenes is an extremely difficult challenge due to large scale variations.
In this paper, we propose a novel dual path multi-scale fusion network architecture with attention mechanism
named SFANet that can perform accurate count estimation as well as present high-resolution density maps for 
highly congested crowd scenes. The proposed SFANet contains two main components: a VGG backbone convolutional 
neural network (CNN) as the front-end feature map extractor and a dual path multi-scale fusion networks
as the back-end to generate density map. 
These dual path multi-scale fusion networks have the same structure, one path is responsible for 
generating attention map by highlighting crowd regions in images, the other path is responsible for fusing
multi-scale features as well as attention map to generate the final high-quality high-resolution density maps.
SFANet can be easily trained in an end-to-end way by dual path joint training. 
We have evaluated our method on four crowd counting datasets 
(ShanghaiTech, UCF\_CC\_50, UCSD and UCF-QRNF). The results demonstrate that with attention mechanism 
and multi-scale feature fusion, the proposed SFANet achieves the best performance on all these datasets 
and generates better quality 
density maps compared with other state-of-the-art approaches.
\end{abstract}

\section{Introduction}
Crowd counting has gained much more attention in recent years because of its various applications such as 
video surveillance, public safety, traffic control. However, due to problems including occlusions, perspective distortions, 
scale variations and diverse crowd distributions, performing precise crowd counting has been a challenging 
problem in computer vision. Some of early methods solve crowd counting problem by detecting each individual pedestrian in a crowd~\cite{ge2009marked,idrees2015detecting,li2008estimating}, while some methods rely on hand-crafted
 features from multi-source~\cite{idrees2013multi}.
These methods may have low performance for the heavily occlusions and diverse crowd distribution scenes. 

\begin{figure}
	\begin{center}
		\includegraphics[width=0.32\linewidth]{IMG_177.jpg}
		\includegraphics[width=0.32\linewidth]{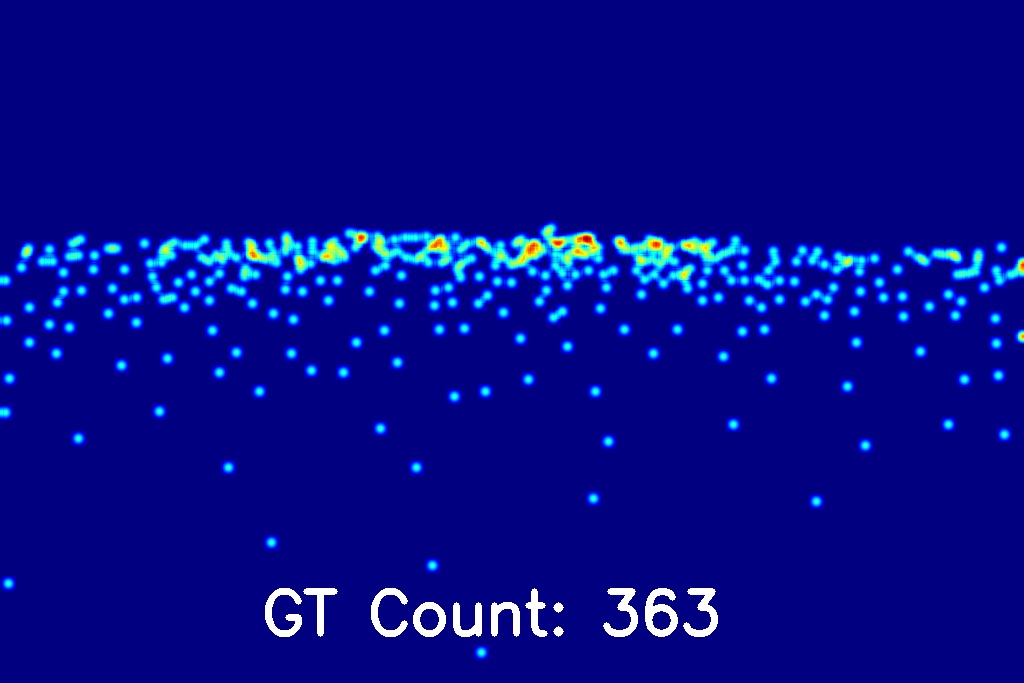}
		\includegraphics[width=0.32\linewidth]{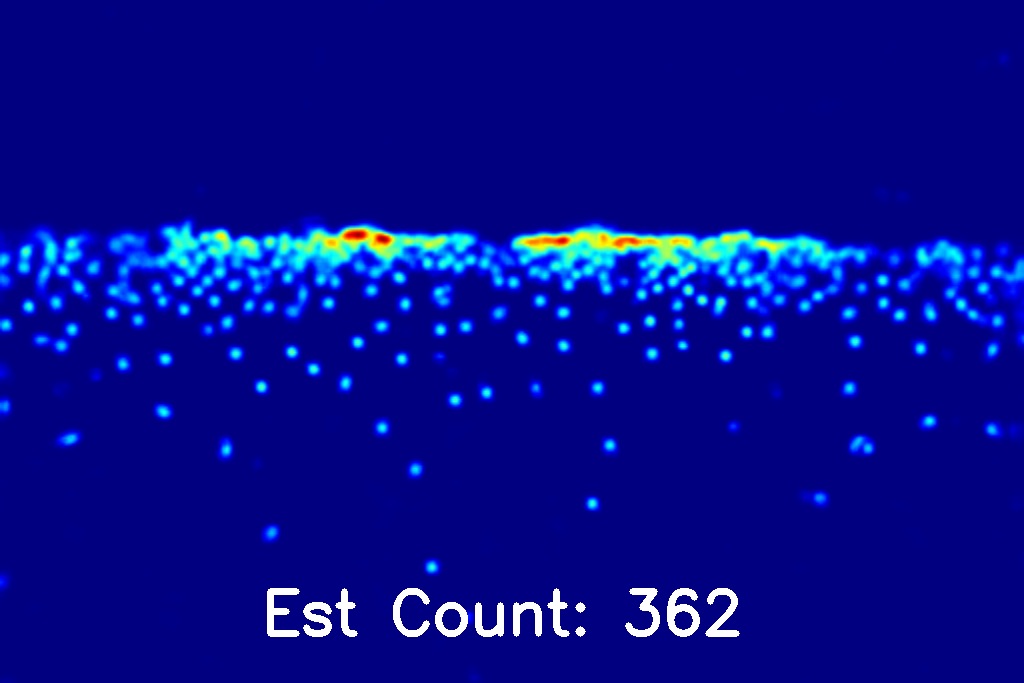}
		\includegraphics[width=0.32\linewidth]{IMG_187.jpg}
		\includegraphics[width=0.32\linewidth]{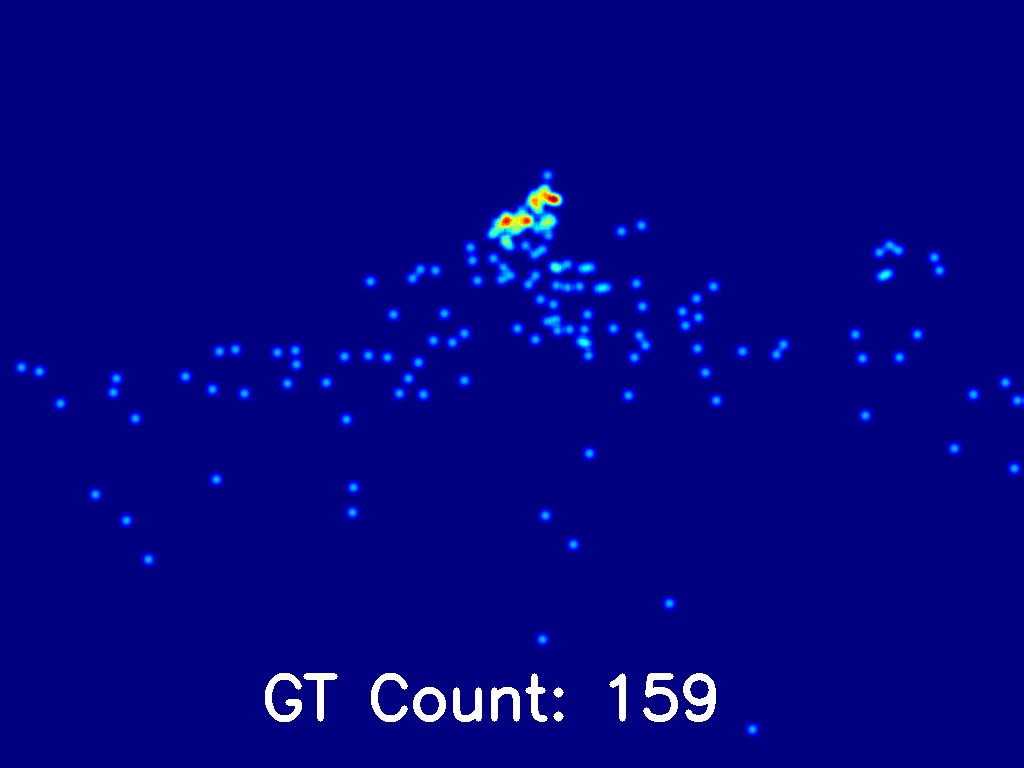}
		\includegraphics[width=0.32\linewidth]{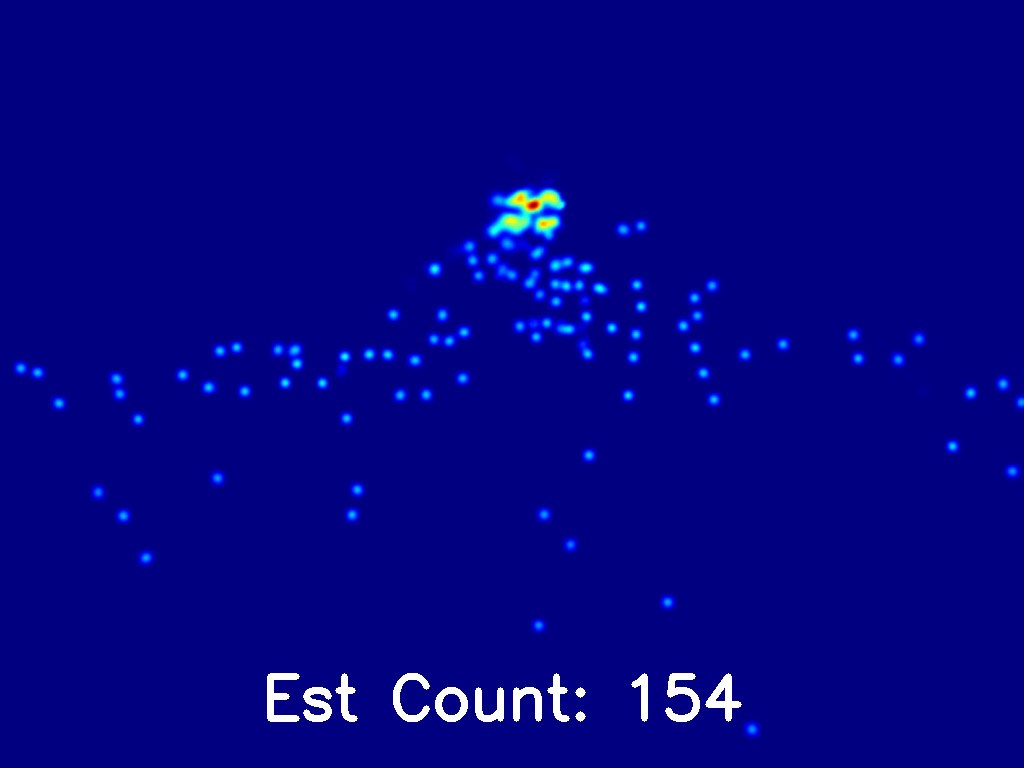}
	\end{center}
	\caption{Top row: large scale variation example. Bottom row: example of diverse crowd distribution with high background noisy.
		From left to right: Input image (from the ShanghaiTech dataset\cite{zhang2016single}), Ground-truth density map, 
		Density map generated by the proposed method.}
	\label{figure1}
\end{figure}

\begin{figure*}[ht]
	\begin{center}
		\includegraphics[width=0.9\linewidth]{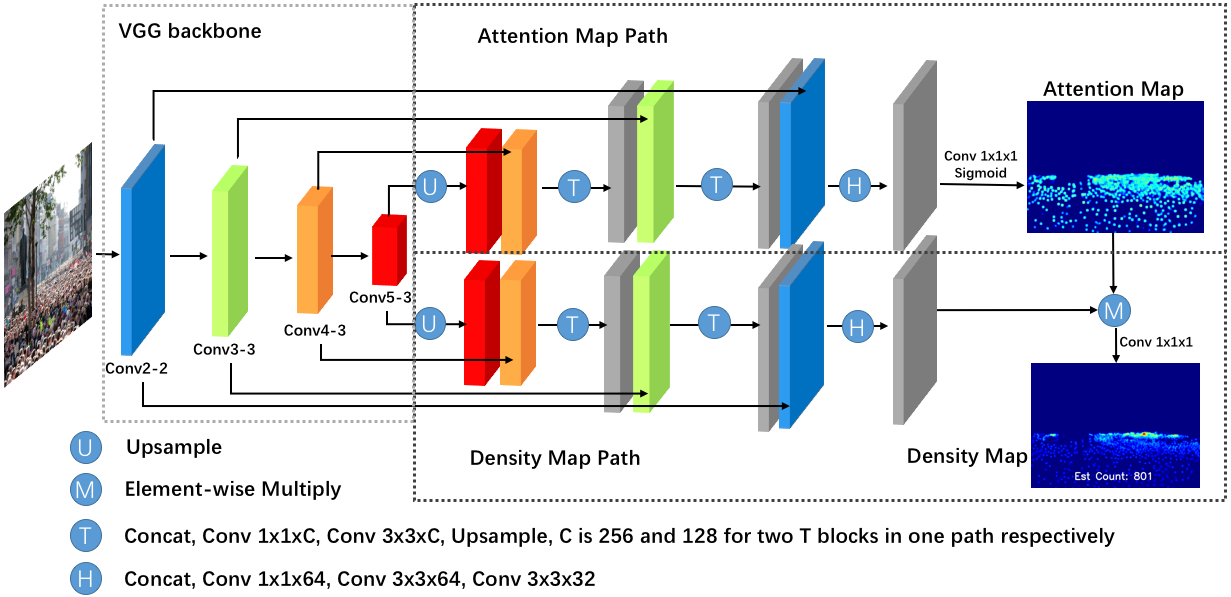}
	\end{center}
	\caption{The architecture of the proposed dual path multi-scale fusion network with attention for crowd counting.}
	\label{figure2}
\end{figure*}

To tackle it, some new methods recently have been developed by utilizing Convolutional Neural Networks (CNNs) for accurate 
crowd density map generation and precise crowd counting. These methods mainly aim to solve two major difficult problems:
large head scale variations caused by camera perspective and diverse crowd distributions with high background noisy scenes.
As shown in Figure~\ref{figure1}, head scale near the camera is much larger than that far from the camera in top row images, 
while bottom row images show diverse crowd distribution with trees and building background noise. 
Certain CNN-based methods deal the issue of 
scale variations via multi-column or multi-resolution network~\cite{zhang2016single,onoro2016towards} to achieve better accuracy. 
Though these methods show improvement in scale variations, they are still restricted by the hand-crafted filters size and 
multi-column structure. In MCNN~\cite{zhang2016single}, each column is dedicated to a certain level of congested scene,
however Li et al. 2018~\cite{li2018csrnet} shows that each column in such branch structure learns nearly identical features.
For the issue of diverse crowd distributions with high background noise scenes, 
MA Hossain et al. 2019~\cite{hossaincrowd} tried to apply attention mechanism to guide 
the network to automatically focus on certain global and local scales appropriate
for the image. But this method only uses the attention model to represent three different
scale levels and still base on multi-columns structure, 
thus it doesn't perform well for congested crowd scenes.

In this paper, we propose a novel method to address
the above-mentioned two major difficult problems of 
precise crowd counting. The network is designed as dual path multi-scale fusion network architecture with attention 
mechanism called SFANet. Figure~\ref{figure2} shows the network architecture used in the
proposed method. Our network adopts the first 13 layers from VGG16-bn as the 
front-end feature map extractor(FME) to extract multi-scale feature maps which contain not only the different level 
semantics information but also the different scale feature information. The low level and small scale 
features can well represent the detail edge patterns which are essential for regressing the value 
of congested region in density map, but don't have sufficient information to 
distinguish head regions from those of tree leaves, buildings, cluttered backgrounds etc. 
On the other hand, the high level and large scale features have useful semantics information
to eliminate the background noise but in lower resolution caused by max-pooling 
operation etc.

Based on the above two points, we design a path of multi-scale feature fusion as density map path (DMP) to 
combine these two advantages of different level features. Another advantage of 
multi-scale feature fusion structure is that we can gain the high-resolution density map by upsample operation.

Another issue is that DMP network does regression for every density map pixel, while 
do not explicitly give
more attention to head regions during training and testing. In the other
word, the training loss, e.g euclidean
distance between ground-truth and estimated density maps, 
may be serious effected by the background noise. To further tackle the high background noise issue, 
we adopt another path of multi-scale feature fusion as attention map path (AMP) with the same structure to 
learn a probability map that indicates high probability head regions.
Then this attention
map is used to suppress non-head regions in the last feature maps of DMP,  
which makes DMP focus on learning the regression task only in high probability head regions.
We also introduce a multi-task loss by adding a attention map loss for AMP, 
which improves the network performance with more explicit supervised signal.

These two  multi-scale feature fusion paths, AMP and DMP co-works in the way like that human being dose
to solve this problem, firstly locate the head regions then count the head numbers.

In summary, our main contributions of this paper are as follows:
\begin{itemize}[topsep=0pt]
	\setlength{\itemsep}{0pt}
	\setlength{\parsep}{0pt}
	\setlength{\parskip}{0pt}
	\item We design a multi-scale fusion network architecture to fuse the feature maps from multi-layers to make the
	network more robust for the head scale variation and
background noise, and also generating high-resolution density maps.
	\item We incorporating the
attention model into the network by adding a path of multi-scale feature fusion as attention map path,
	which makes the proposed method focus on head regions for the density map regression task, therefore
	improving its robustness to complex backgrounds and diverse crowd distributions.
	\item We propose a novel multi-task training loss, combining Euclidean loss and attention map loss to make 
	network convergence faster and better performance. The former loss minimizes the pixel-wise error and 
	the latter one focus on locating the head regions.
\end{itemize}

\section{Related work}
\textbf{Traditional counting methods: }Traditional counting methods for crowd counting relied on hand-crafted representations to extract low
level features. These methods can be categorized into two: detection-based methods and
regression-based methods.

Detection-based methods estimate
the number of people using sliding
window based detection algorithms\cite{dollar2012pedestrian} and hand
crafted features extracted from heads or bodys with low-level descriptors such as
Haar
wavelets\cite{viola2004robust} and HOG\cite{dalal2005histograms}. However severe
 occlusions make these methods perform poorly in congested scenes.

To overcome the problem of occlusions, regression-based methods 
aim to learn the mapping between
features extracted from cropped image patches to their count or density\cite{ryan2009crowd}. 
The extracted features are used to generate low-level information, 
which is leaned by a regression
model. Instead of directly regressing the 
total crowd count, Lempitsky et al.\cite{lempitsky2010learning} propose a
method to solve counting problem by modeling a linear
mapping between features in the image region and its density maps.
Pham et al.\cite{pham2015count} observed the difficulty of learning a linear mapping 
and proposed method to learn a non-linear mapping using a random forest regression.

\textbf{CNN-based counting methods: }Recently CNN-based methods have shown a great success in crowd counting 
and density estimation. Walach
et al.\cite{walach2016learning} used CNNs with boosting and selective sampling.
Zhang et al.\cite{zhang2015cross} propose a deep convolutional neural network for crowd counting 
with two related learning objectives, crowd density and crowd count.
Different from the existing patch-based estimation methods, Shang
et al.\cite{shang2016end} used a network that simultaneously predict local and global counts
for whole input images by taking advantages of contextual information. 
Boominathan et al.\cite{boominathan2016crowdnet}
use
a dual-column network combining shallow and deep layers to generate density maps.
Zhang et al.\cite{zhang2016single} designed multi-column CNN
(MCNN) to tackle the large scale variation in crowd scenes.
 With similar idea,
Onoro et al.\cite{onoro2016towards} proposed a scale-aware network, 
called Hydra, to training the network with a
pyramid of image patches at multiple scales.
Based on multi-scale CNN architecture, Sam et al.\cite{sam2017switching} 
proposed switch CNN by training a 
switch classifier to select the best CNN regressor for image patches.
Sindagi et al.\cite{sindagi2017generating} proposed contextual pyramid network for generating high-quality crowd density
by explicitly incorporating global and local contextual information.

More recently, Li et al.\cite{li2018csrnet} proposed CSRNet by using the dilated convolutional layers 
to aggregate the multi-scale contextual information in the congested scenes.
Zhang et al.\cite{zhang2018crowd} proposed a scale-adaptive CNN(SaCNN) 
concatenating multiple feature maps
of different scales with a VGG backbone.
Cao et al.\cite{cao2018scale} presented encoder-decoder network for scale aggregation. 
The encoder extracts multi-scale features with scale
aggregation modules and the decoder generates high-resolution density
maps by using a set of transposed convolutions. 

\textbf{Counting methods with Attention: } Recently, with attention model 
widely used for various computer vision tasks, such as image classification, 
segmentation and object detection, some researchers attempted to use the method in crowd counting. Liu et al.\cite{liu2018decidenet} proposed framework named DecideNet to estimate crowd counts via adaptively adopting detection and regression based count estimations under the guidance from the attention mechanism. However, training
both detection and regression needs large amount of computation.
MA Hossain et al.~\cite{hossaincrowd} proposed a scale-aware attention network 
by combining different global and local scale levels. Because the attention model is only applied
to three different scale level column branches, it can't handle large scale variation in
complex crowd scene.

\section{Proposed Approach(SFANet)}
Inspired by the success use of feature pyramid networks and attention mechanism, 
we proposed SFANet consist of VGG16-bn backbone feature map extractor(FME)
and dual path multi-scale fusion networks with attention(DMP and AMP). 
Input images are first feed into FME to extract 
multi-scale features. The density map path(DMP) use concatenate and upample
to fuse multi-scale features, while the attention map path(AMP) incorporates
attention model to emphasize head regions to tackle background noise and non-uniformity of crowd distributions. In addition, a attention map loss is
introduced to compensate Euclidean loss with more explicit supervised information.
The architecture of the proposed SFANet is illustrated in Fig.\ref{figure2}and discussed in detail as follows.

\subsection{SFANet architecture}
\textbf{Feature map extractor(FME)}: Most previous works using the multi-column
architecture with different filter sizes to deal with the large scale variation due to perspective
effect. We instead use a single backbone network with
a single filter size as the feature map extractor. We adopt all 3*3 filters in the network, which require far less computation than large filters and can build 
deeper network. We choose a pre-trained VGG16 with batch 
normalization as the frond-end
feature map extractor due to its strong feature represent ability and easily to be 
concatenated by the back-end dual path networks. The first 13 layers from conv1-1
to conv5-3 are involved to output feature maps with sizes 1/2, 
1/4, 1/8 and 1/16 of the original 
input size. Four source layers, conv2-2, conv3-3, conv4-3 and conv5-3,
which represent multi-scale features and multi-level semantic information,
will be concatenated by both DMP and AMP.

\begin{figure}[ht]
	\begin{center}
		\includegraphics[width=0.9\linewidth]{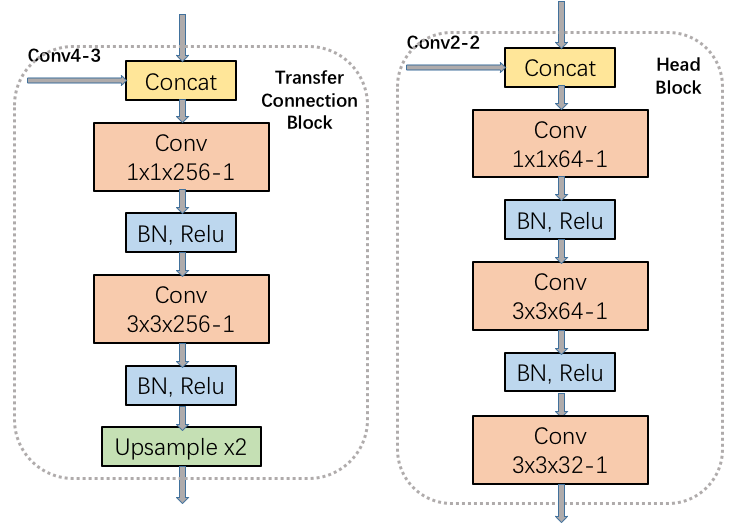}
	\end{center}
	\caption{The structure of the proposed Transfer Connection Block and Head Block. A convolutional layer is denoted as
Conv (kernel size) $\times$ (kernel size) $\times$ (number of channels)-stride. BN, Relu represent for standard batch normalization and Relu layers.}
	\label{figure3}
\end{figure}

\textbf{Density map path(DMP)}: The DMP of SFANet is constructed in 
feature pyramid structure as illustrated in Fig.\ref{figure2}. Conv5-3 feature maps
firstly is upsampled by factor 2, and then concatenate feature maps of 
conv4-3. The detail of transfer connection block T is shown as Fig.\ref{figure3}, which contains 
concat, conv1$\times$1$\times$256, conv3$\times$3$\times$256 
and upsample sub-layers. The 
second T block has the similar structure concatenating conv3-3 
with only different channel size 128,
that is  concat, conv1$\times$1$\times$128, conv3$\times$3$\times$128 
and upsample. Then concatenated outputs of second T block and conv2-2 are
feed into header block H with concat, conv1$\times$1$\times$64, conv3$\times$3$\times$64 and conv3$\times$3$\times$32 shown as Fig.\ref{figure3}.
Every 1 $\times$ 1
convolution before the 3 $\times$ 3 is used
to reduce the
computational complexity. Due to previous three upsample layers,
we can retrieve the final high resolution feature maps with 1/2 size of the original 
input. Then element-wise multiple is applied on attention map and the last
density feature maps to generate refined density feature maps $F_{refine}$ as
equation \ref{eq1}:
\begin{equation}\label{eq1}
F_{refine} = f_{den} \otimes M_{Att}
\end{equation}
where $f_{den}$ is the last density features, $M_{Att}$ is attention map, 
$\otimes$ denotes element-wise multiply. Before this operation, 
$M_{Att}$ is expanded as the same channel as $f_{den}$.
At last, we use a simple convolution with kernel 1$\times$1$\times$1 
to generate the high-quality density map $M_{den}$. 
Batch normalization is applied after every convolutional layer because we find that batch training and batch normalization can stabilize the training process and accelerate loss
convergence. We also apply Relu after every convolutional layer except
the last one.

\begin{figure*}[ht]
	\centering
	\begin{subfigure}[t]{0.23\linewidth}
		\centering
		\includegraphics[width=1\linewidth]{IMG_177.jpg}
	\end{subfigure}
	\begin{subfigure}[t]{0.23\linewidth}
		\centering
		\includegraphics[width=1\linewidth]{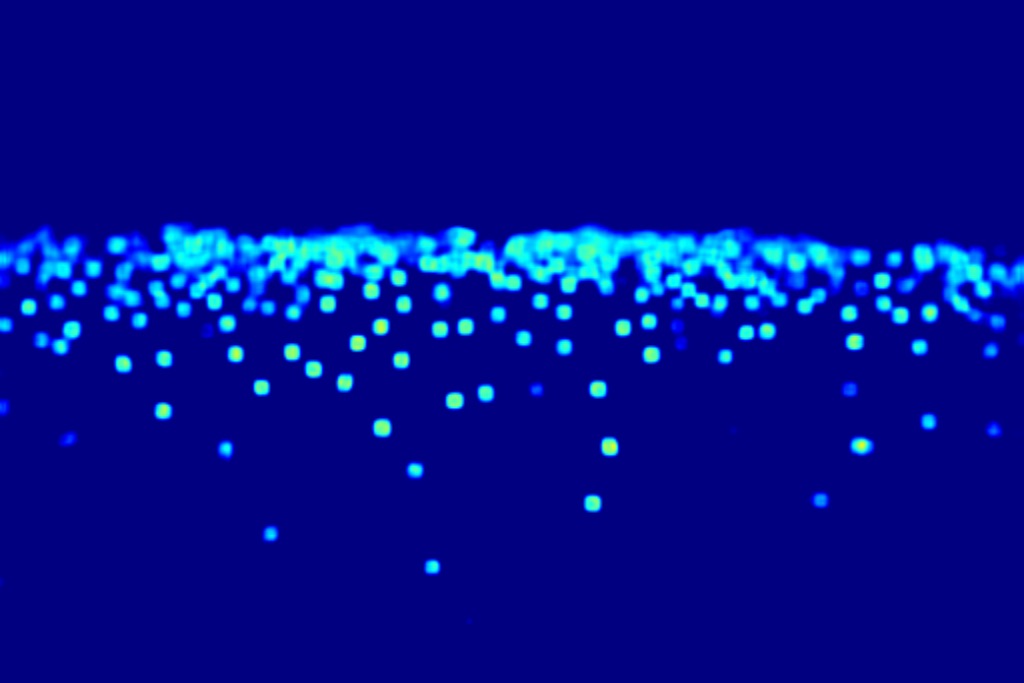}
	\end{subfigure}
	\begin{subfigure}[t]{0.23\linewidth}
		\centering
		\includegraphics[width=1\linewidth]{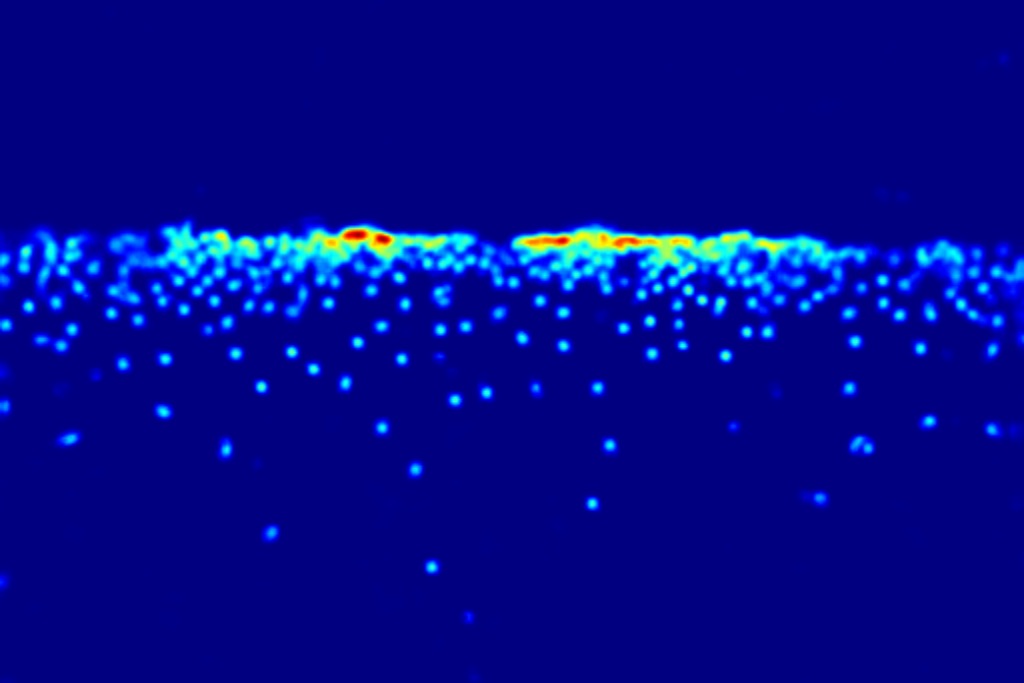}
	\end{subfigure}
	\begin{subfigure}[t]{0.23\linewidth}
		\centering
		\includegraphics[width=1\linewidth]{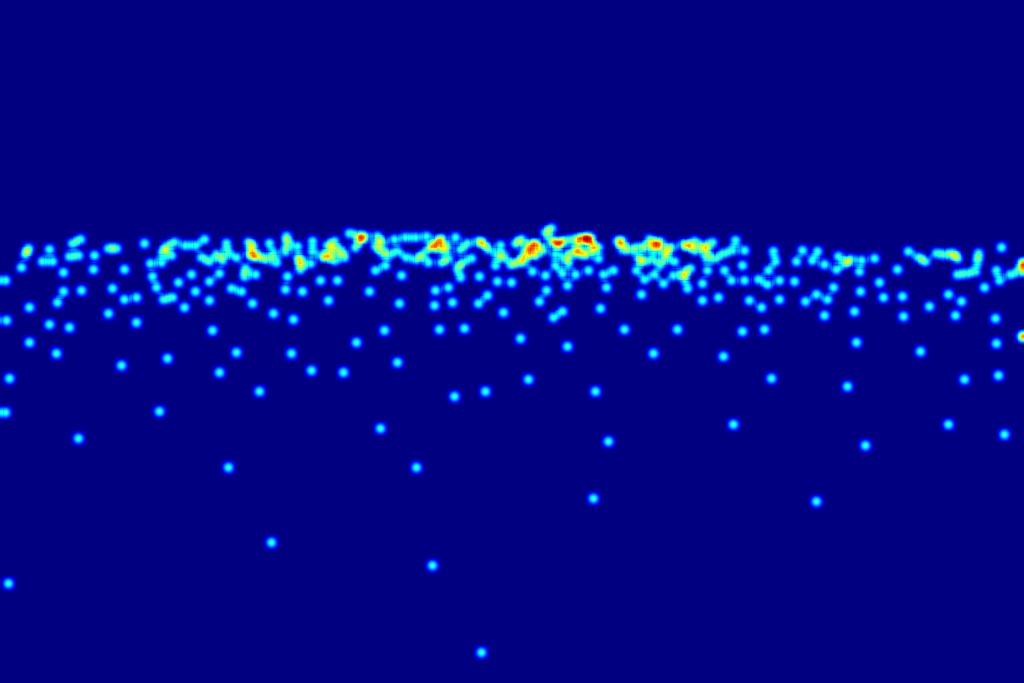}
	\end{subfigure}
	\begin{subfigure}[t]{0.23\linewidth}
		\centering
		\includegraphics[width=1\linewidth]{IMG_157.jpg}
	\end{subfigure}
	\begin{subfigure}[t]{0.23\linewidth}
		\centering
		\includegraphics[width=1\linewidth]{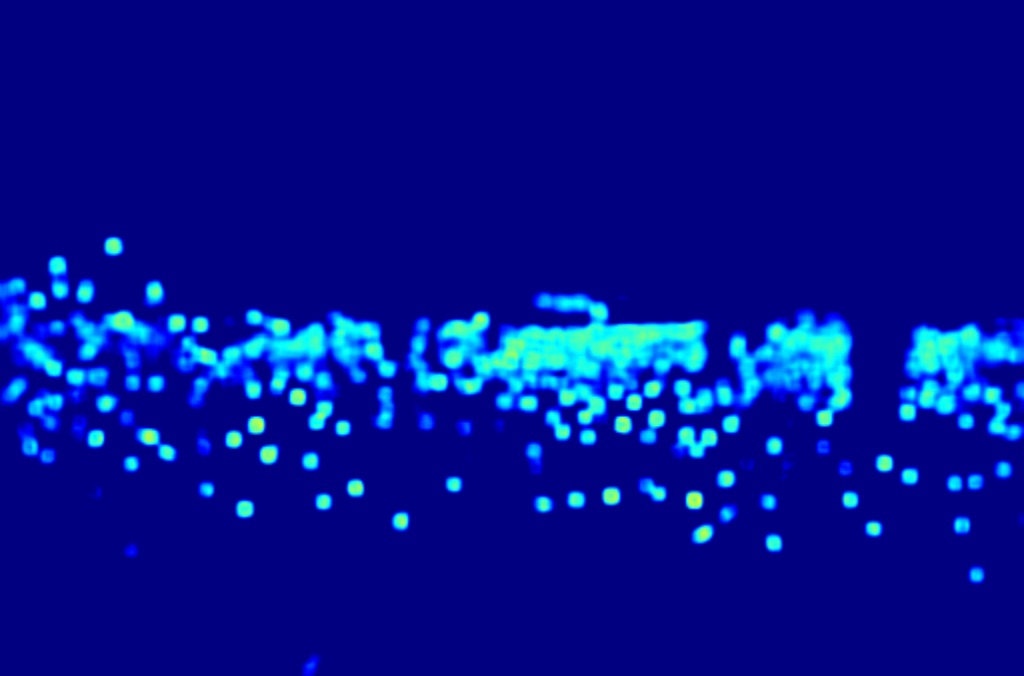}
	\end{subfigure}
	\begin{subfigure}[t]{0.23\linewidth}
		\centering
		\includegraphics[width=1\linewidth]{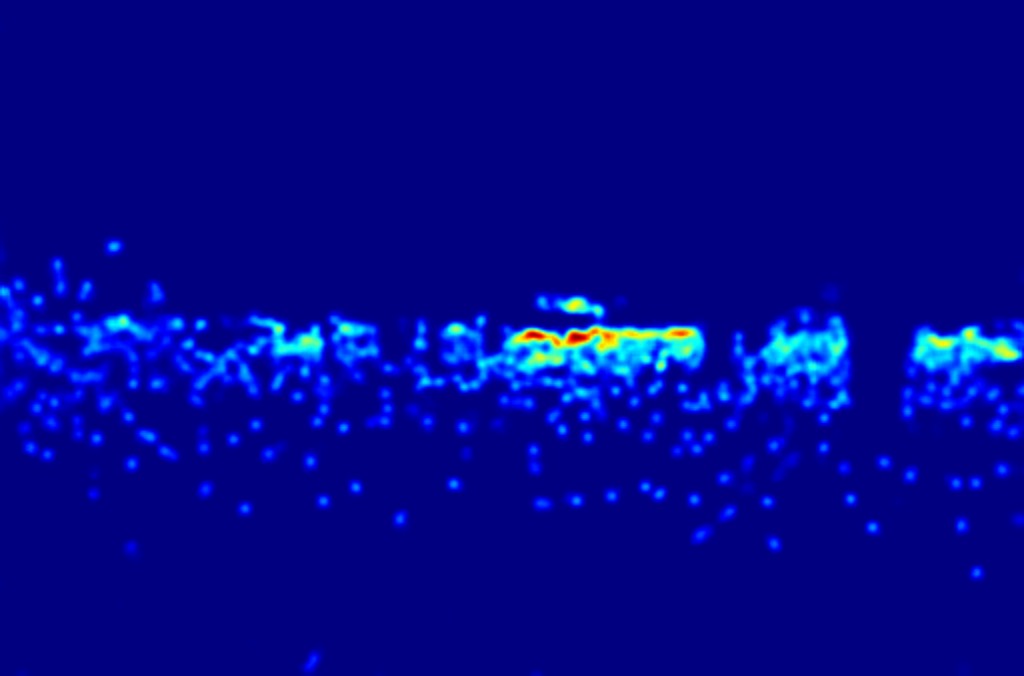}
	\end{subfigure}
	\begin{subfigure}[t]{0.23\linewidth}
		\centering
		\includegraphics[width=1\linewidth]{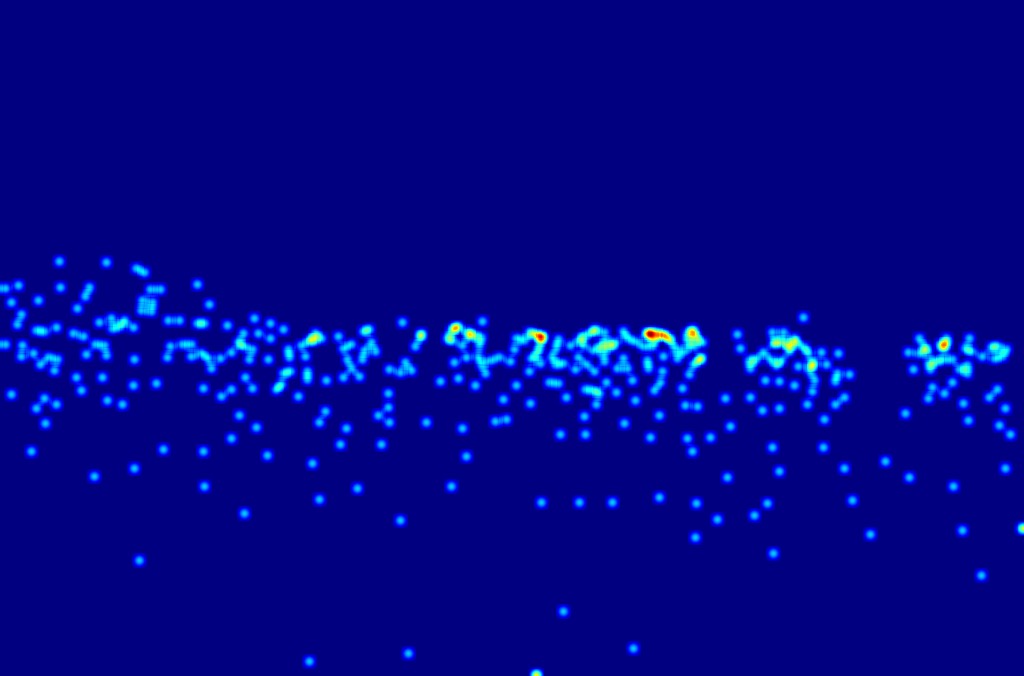}
	\end{subfigure}
	\begin{subfigure}[t]{0.23\linewidth}
		\centering
		\includegraphics[width=1\linewidth]{IMG_87.jpg}
		\caption{Input Image}\label{fig:1a}		
	\end{subfigure}
	\begin{subfigure}[t]{0.23\linewidth}
		\centering
		\includegraphics[width=1\linewidth]{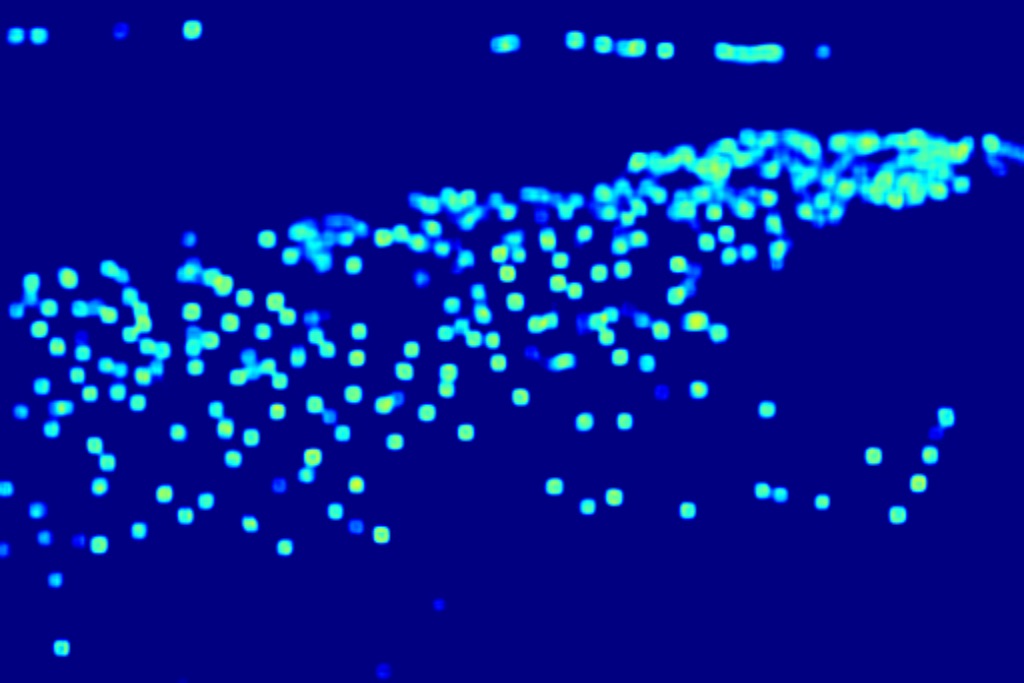}
		\caption{Attention Map}\label{fig:1b}
	\end{subfigure}
	\begin{subfigure}[t]{0.23\linewidth}
		\centering
		\includegraphics[width=1\linewidth]{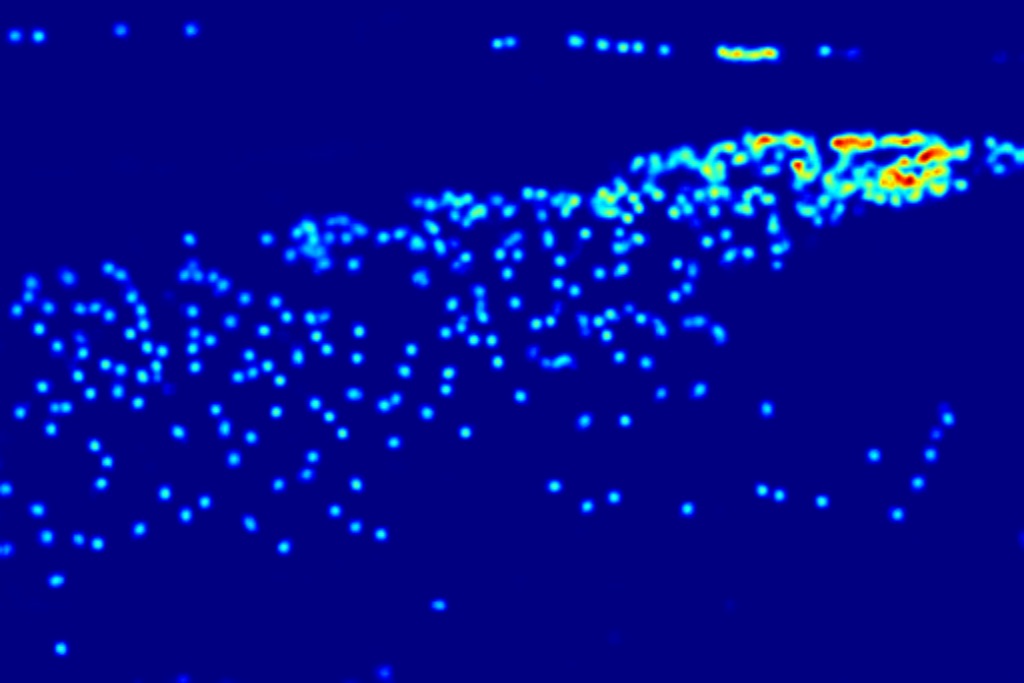}
		\caption{Density Map}\label{fig:1c}
	\end{subfigure}
	\begin{subfigure}[t]{0.23\linewidth}
		\centering
		\includegraphics[width=1\linewidth]{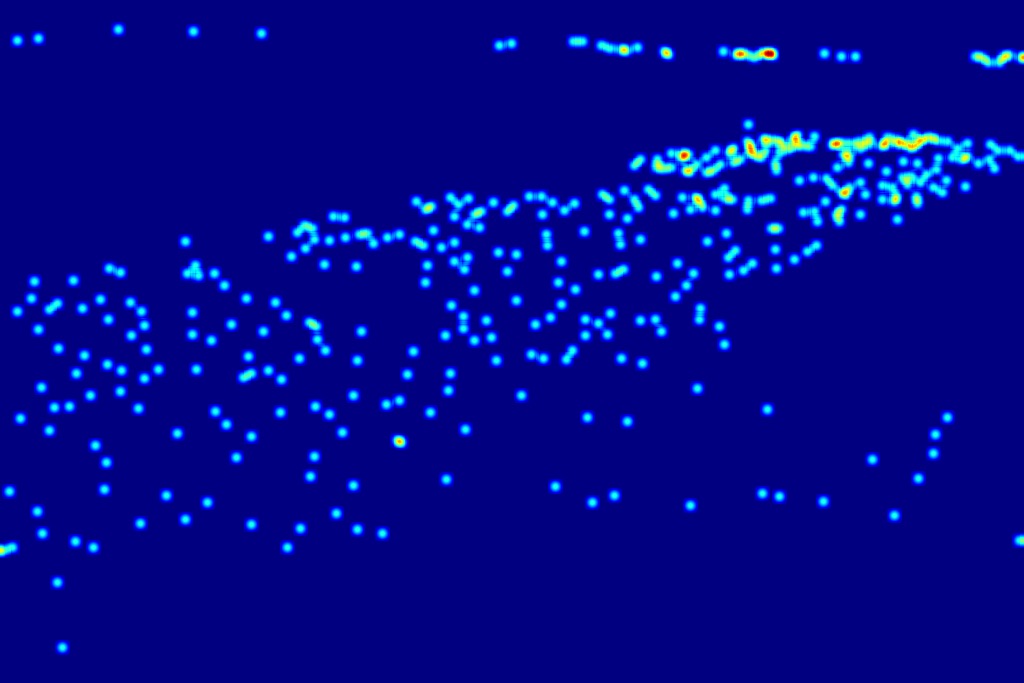}
		\caption{Ground Truth}\label{fig:1d}
	\end{subfigure}
	\caption{Attention maps and density maps of ShanghaiTech dataset generated by SFANet.}\label{figure4}
\end{figure*}

\textbf{Attention map path(AMP)}: The AMP of SFANet has the similar 
structure with DMP, and output probability map to indicate head probability
in each pixel of the density feature map. In this work, we introduce the
attention model as follows. Suppose convolutional
features output by head block as $f_{att}$, the attention map $M_{Att}$ 
is generated as:
\begin{equation}\label{eq2}
M_{Att} = Sigmoid(W \textcircled{c} f_{att} + b)
\end{equation} 
where $W$, $b$ is the $1 \times 1 \times 1 $ convolution layer weights and bias, \textcircled{c} 
denotes the convolution operation and $Sigmoid$ denotes the sigmoid activation function. The sigmoid activation function gives out $(0, 1)$ probability scores
to make network discriminate head location and background. The visualization
of $M_{att}$ can be seen in Fig.\ref{figure4}. The proposed attention map loss
will be further discussed the next section.

\subsection{Loss Function}
Most of previous methods use Euclidean distance as the loss
function for crowd counting. The Euclidean loss is used to measure estimation error at pixel
level, which is defined as follows:
\begin{equation}\label{eq3}
L_{den} = \frac{1}{N}\sum_{i=1}^{N}\left \| F(X_i;\Theta) - D_i^{GT} \right \|^2
\end{equation}
where $F(X_i;\Theta)$ is the estimated density map. $\Theta$ is a set of learnable 
parameters in the proposed
network. $X_i$ is the input image and $D_i^{GT}$ is
the ground truth density map. $N$ is the number of train batch.

Besides the density map regression, we introduce
another attention map loss function in attention map path training process. The
attention map loss function is a binary class entropy, defined as equation\ref{eq4}:
\begin{equation}\label{eq4}
L_{att} = -\frac{1}{N}\sum_{i=1}^{N}(A_i^{GT}\log(P_i) + (1-A_i^{GT})\log(1-P_i))
\end{equation}
where $A_i^{GT}$ is the attention map groundtruth. $P_i$ is the probability of each pixel in 
predict attention map activated by sigmoid function.

The entire network is trained using the following
unified loss function:
\begin{equation}\label{eq5}
L = L_{den} + \alpha L_{att}
\end{equation}
where $\alpha$ is a weighting weight that is set as $0.1$ in the experiments.
We use this multi-task combine loss to do end-to-end dual path joint training.

\section{Training method}
In this section, we illustrate details of SFANet
end-to-end training method.

\subsection{Density map groundtruth}
To obtain the ground-truth density maps $D^{GT}$, we follow the method of generating density maps in
\cite{sindagi2017generating} by using the same size of Gaussian kernels for all objects.
Supposing there is a point at pixel
$x_i$ that denote the position of pedestrian head in the scene,
the
corresponding ground-truth density map $D^{GT}$ can be computed by blurring each head annotation
using a Gaussian kernel. The generation
of density map $D^{GT}$ can be formulated as:
\begin{equation}\label{eq6}
D^{GT} = \sum_{i=1}^{C}\delta(x-x_i)\times G_{\mu,\rho^2}(x)
\end{equation}
For each annotation target head $x_i$ in the ground truth $\delta$, we convolve $\delta(x-x_i)$ by
a Gaussian kernel $ G_{\mu,\rho^2}$ with parameter $\mu$ (kernel size) and $\rho$ (standard deviation), 
where $x$ is the position of pixel in the image, $C$ is number of head annotations.
In experiment, we set $\mu = 15$ and $\rho=4$ for all datasets
except UCF-QRNF. Due to the large image size in UCF-QRNF dataset, we resize the image to $1024\times768$
for both training and testing. So we firstly use a adaptive Gaussian kernel with $\mu = 1+15 \times w / 1024 // 2 \times 2$
and $\rho=(\mu+4)/4$, where $w$ is the image width, then resize density map groundtruth 
to the same size as image resized.

\subsection{Attention map groundtruth}
Based on density map groundtruth, we continue use Gaussian kernel to compute attention map groundtruth as follows:
\begin{equation}\label{eq7}
\mathbb{Z}= D_i^{GT}\times G_{\mu,\rho^2}(x)
\end{equation}
\begin{equation}\label{eq8}
\forall x \in \mathbb{Z}, A_i^{GT}(x) = \left\{
\begin{array}{lr}
0,\ \ x < th \\
1,\ \ x \geq th
\end{array}
\right.
\end{equation}
where $th$ is the threshold set as $0.001$ in our experiments. With equation\ref{eq7}, \ref{eq8}, we obtain a binary attention map groundtruth in order to
guide the AMP to focus on the head regions and also the surround places. In experiment, we set 
$\mu = 3$ and $\rho=2$ for generating attention map groundtruth. 

\subsection{Training details}
In training procedure, we firstly resize the short side of image to 512 if the short side is less
than 512, followed by a random 
scale change with ratio $[0.8,1.2]$. Images patches with fixed size $(400\times400)$ are cropped at random
locations, then they are randomly horizontal flipped with probability $0.5$
and processed by gamma contrast transform 
using parameter $[0.5,1.5]$ with probability $0.3$ for data augmentation. 
For those datasets with gray images eg. ShanghaiTech A, 
we also randomly change the color images to gray with probability $0.1$.
As discussion in above section, we resize the image to fixed $1024\times768$ before any data augment for
UCF-QRNF dataset.
In order to match the output size of SFANet, density map and
attention map groundtruth are both resized to half resolution of input image patches.

First 13 layers of pre-trained VGG16 with batch normalization is applied as the front-end feature 
extractor. The rest network parameters are
randomly initialized by a Gaussian distributions with mean $0$ and standard
deviation of $0.01$. Adam\cite{kingma2014adam} optimizer with learning rate of $1e-4$ and weight decay of 
$5e-3$ is used
to train the model, because it shows faster convergence than standard stochastic
gradient descent with momentum in our experiments. We use batch size 30 in training procedure,
which stabilizes the training loss change.

\section{Experiments}
In this section, we present the experimental details
and evaluation results on 4 public challenging datasets:
ShanghaiTech\cite{zhang2016single}, UCF\_CC\_50\cite{idrees2013multi}, 
UCF-QRNF\cite{idrees2018composition} and UCSD\cite{chan2008privacy}. 
We evaluate the performance via the mean absolute
error (MAE) and mean square error (MSE) commonly. These metrics
are defined as follows:
\begin{equation}\label{eq9}
\setlength{\abovedisplayshortskip}{0cm} 
MAE = \frac{1}{N}\sum_{i=1}^{N}\left| C_i - C_i^{GT}\right|
\setlength{\belowdisplayshortskip}{0cm}
\end{equation}
\begin{equation}\label{eq10}
\setlength{\abovedisplayshortskip}{0cm} 
MSE = \sqrt{\frac{1}{N}\sum_{i=1}^{N}\left| C_i - C_i^{GT}\right|^2}
\end{equation}
where $N$ is the number of test image set, $C_i$ and $C_i^{GT}$ are the estimated count of people 
and the ground truth of counting respectively.

\subsection{ShanghaiTech dataset}
\begin{figure*}[ht]
	\centering
	\begin{subfigure}[t]{0.23\linewidth}
		\centering
		\includegraphics[width=1\linewidth]{IMG_11.jpg}
	\end{subfigure}
	\begin{subfigure}[t]{0.23\linewidth}
		\centering
		\includegraphics[width=1\linewidth]{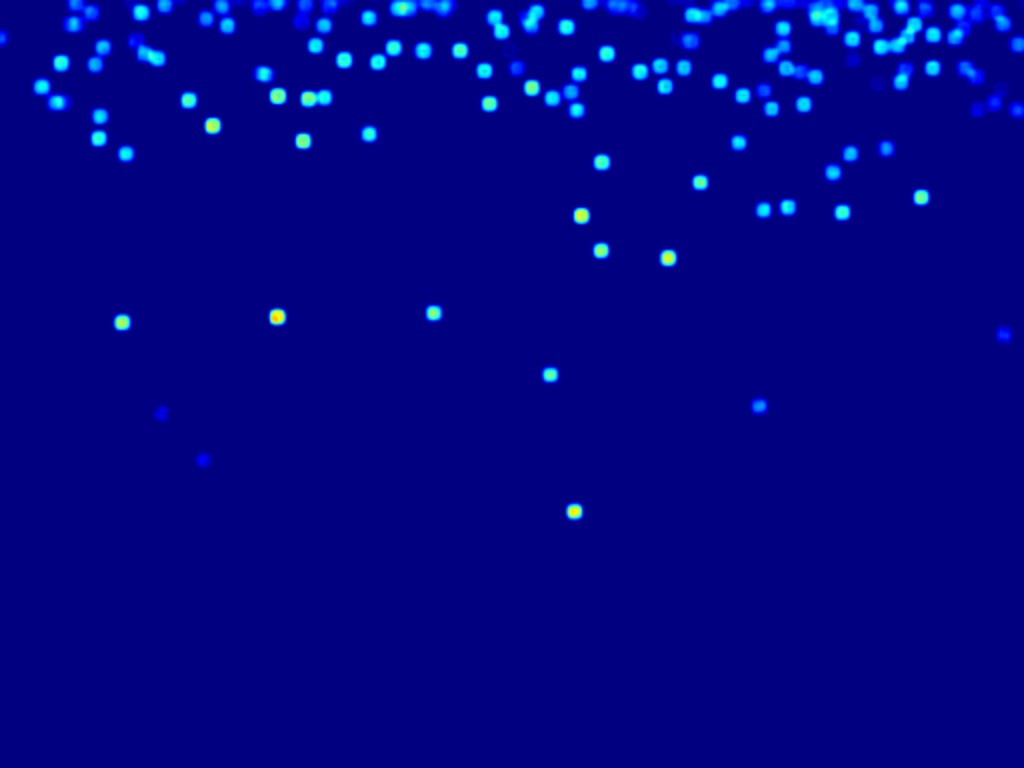}
	\end{subfigure}
	\begin{subfigure}[t]{0.23\linewidth}
		\centering
		\includegraphics[width=1\linewidth]{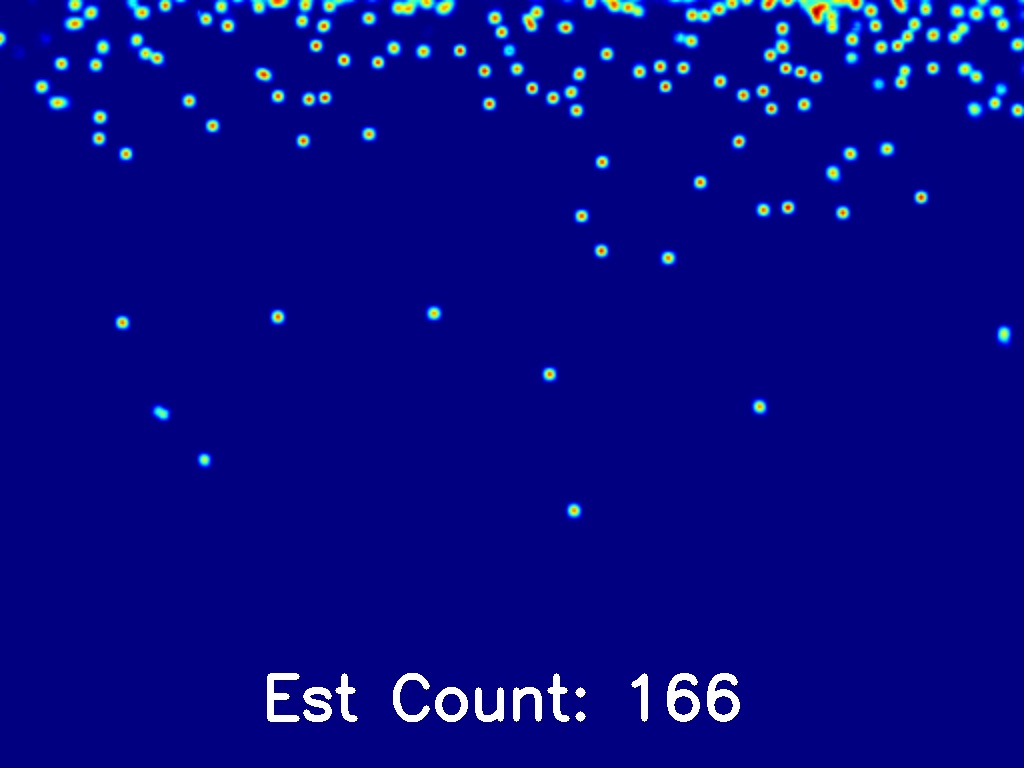}
	\end{subfigure}
	\begin{subfigure}[t]{0.23\linewidth}
		\centering
		\includegraphics[width=1\linewidth]{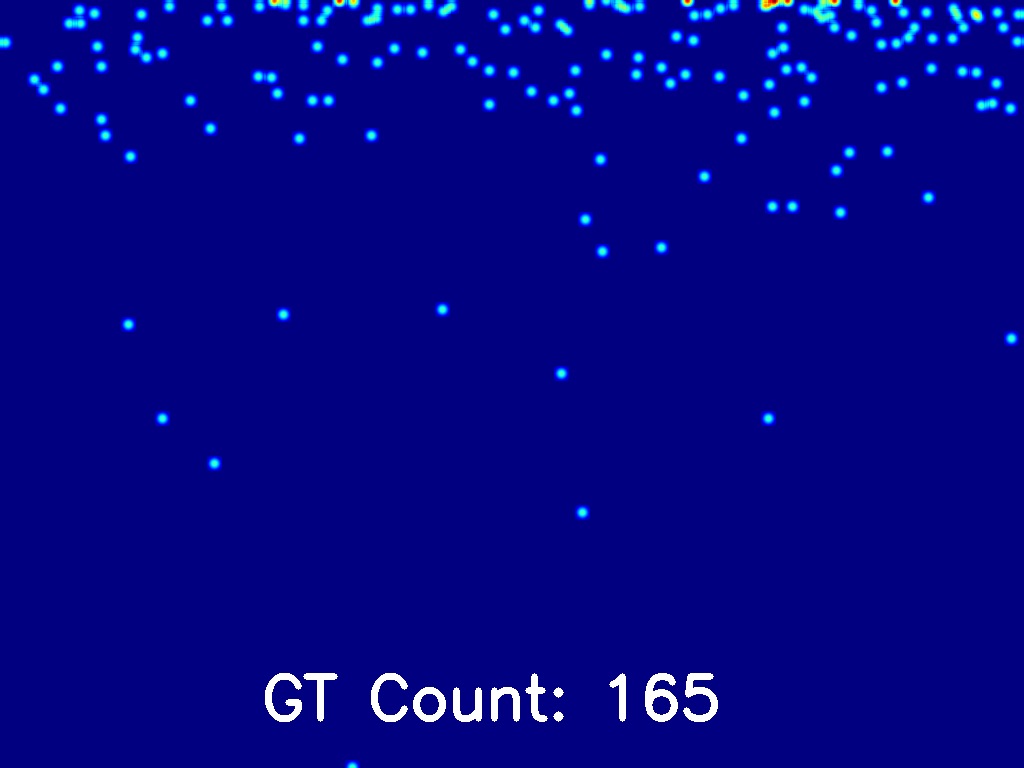}
	\end{subfigure}
	\begin{subfigure}[t]{0.23\linewidth}
	    \centering
	    \includegraphics[width=1\linewidth]{IMG_213.jpg}
    \end{subfigure}
	\begin{subfigure}[t]{0.23\linewidth}
		\centering
		\includegraphics[width=1\linewidth]{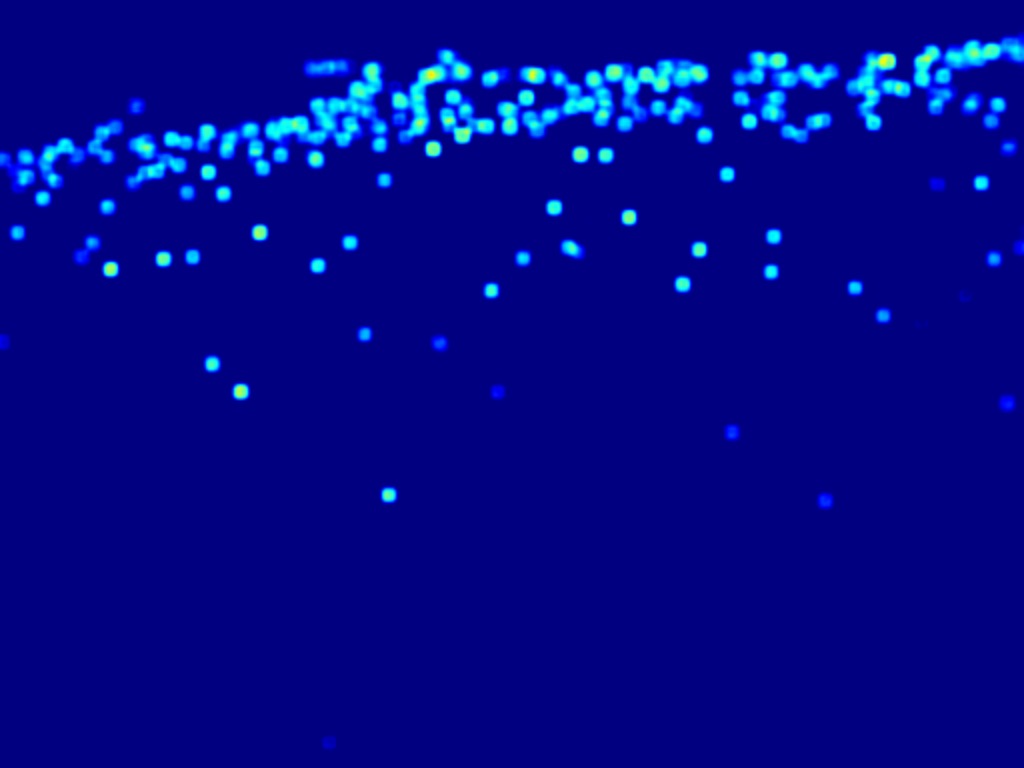}
	\end{subfigure}
	\begin{subfigure}[t]{0.23\linewidth}
		\centering
		\includegraphics[width=1\linewidth]{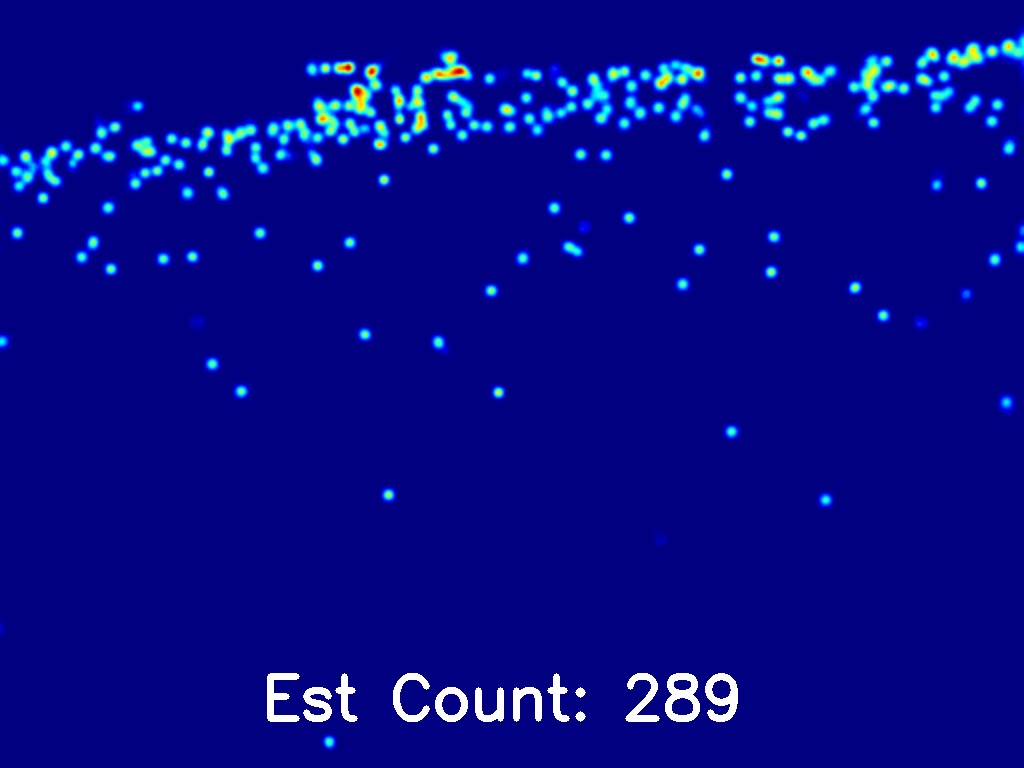}
	\end{subfigure}
	\begin{subfigure}[t]{0.23\linewidth}
		\centering
		\includegraphics[width=1\linewidth]{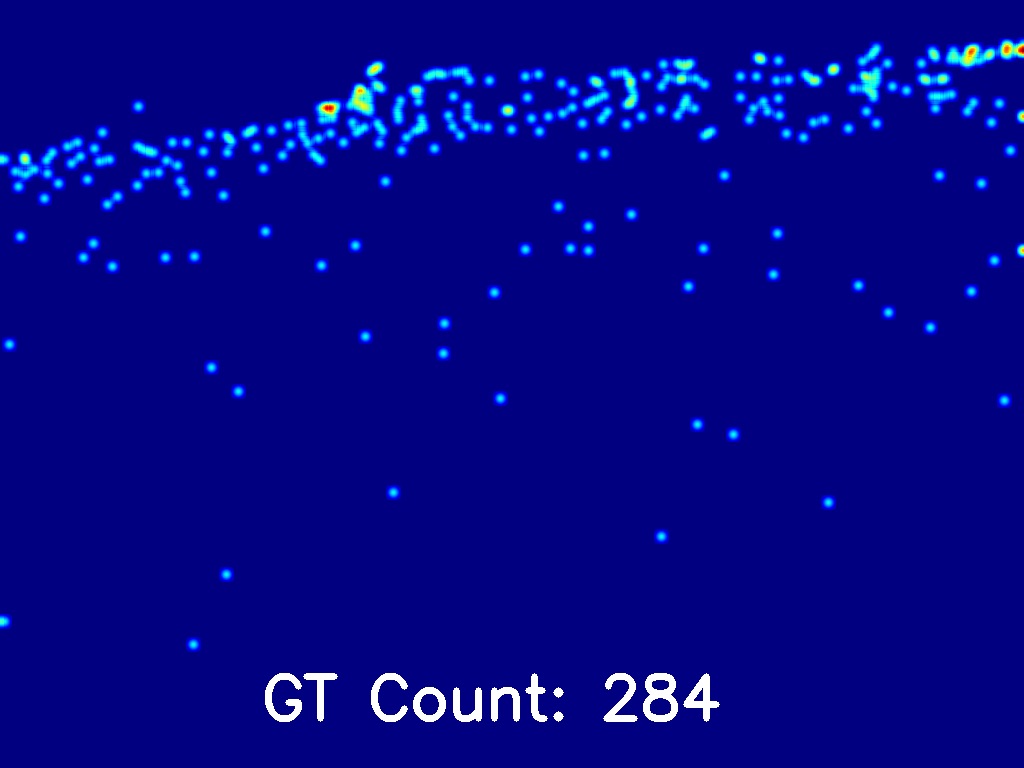}
	\end{subfigure}
	\begin{subfigure}[t]{0.23\linewidth}
		\centering
		\includegraphics[width=1\linewidth]{IMG_32.jpg}
	\end{subfigure}
	\begin{subfigure}[t]{0.23\linewidth}
		\centering
		\includegraphics[width=1\linewidth]{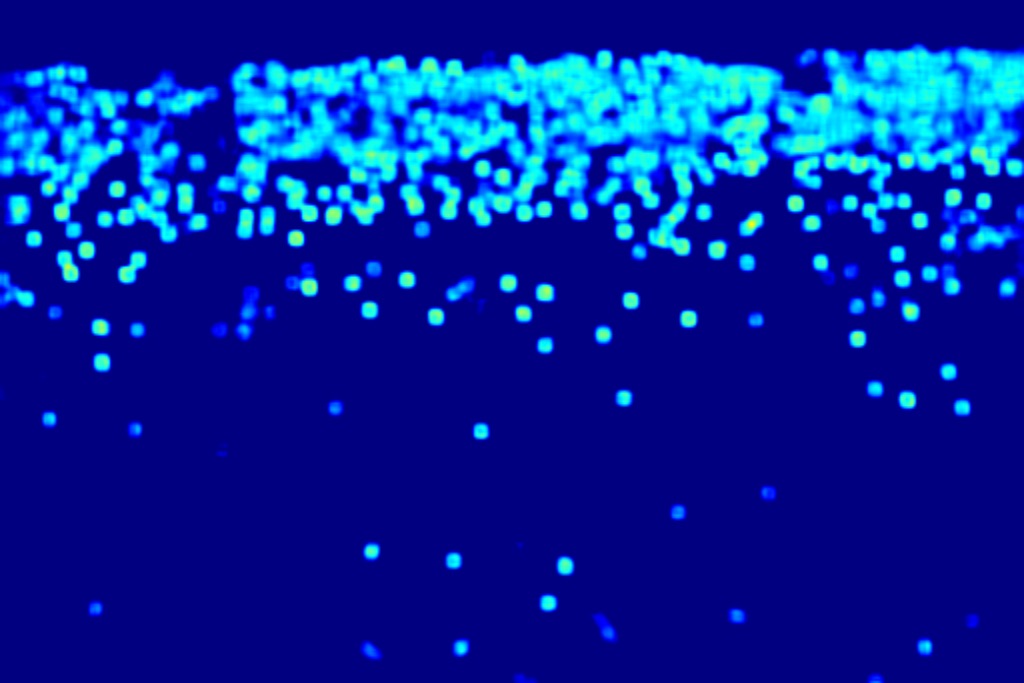}
	\end{subfigure}
	\begin{subfigure}[t]{0.23\linewidth}
		\centering
		\includegraphics[width=1\linewidth]{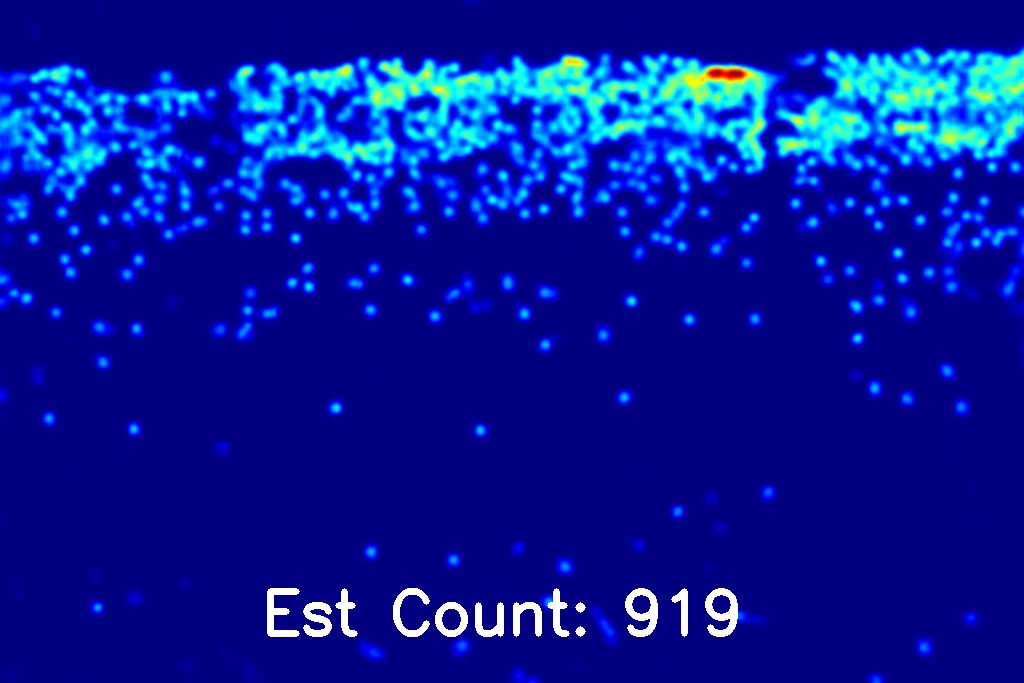}
	\end{subfigure}
	\begin{subfigure}[t]{0.23\linewidth}
		\centering
		\includegraphics[width=1\linewidth]{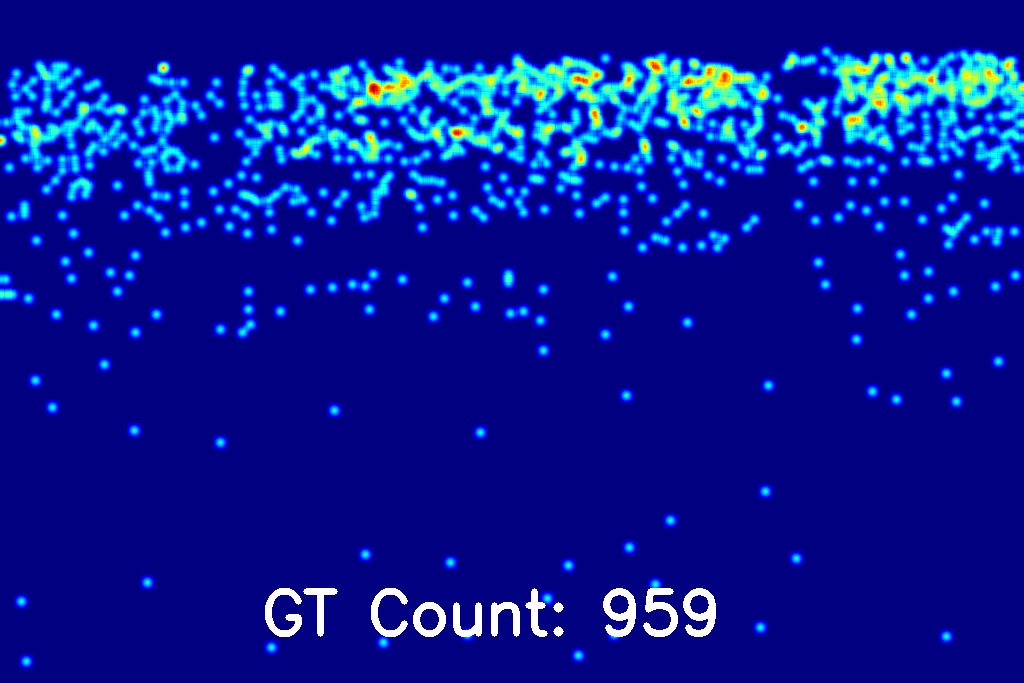}
	\end{subfigure}
	\begin{subfigure}[t]{0.23\linewidth}
		\centering
		\includegraphics[width=1\linewidth]{a_IMG_11.jpg}
		\caption{Input Image}
	\end{subfigure}
	\begin{subfigure}[t]{0.23\linewidth}
		\centering
		\includegraphics[width=1\linewidth]{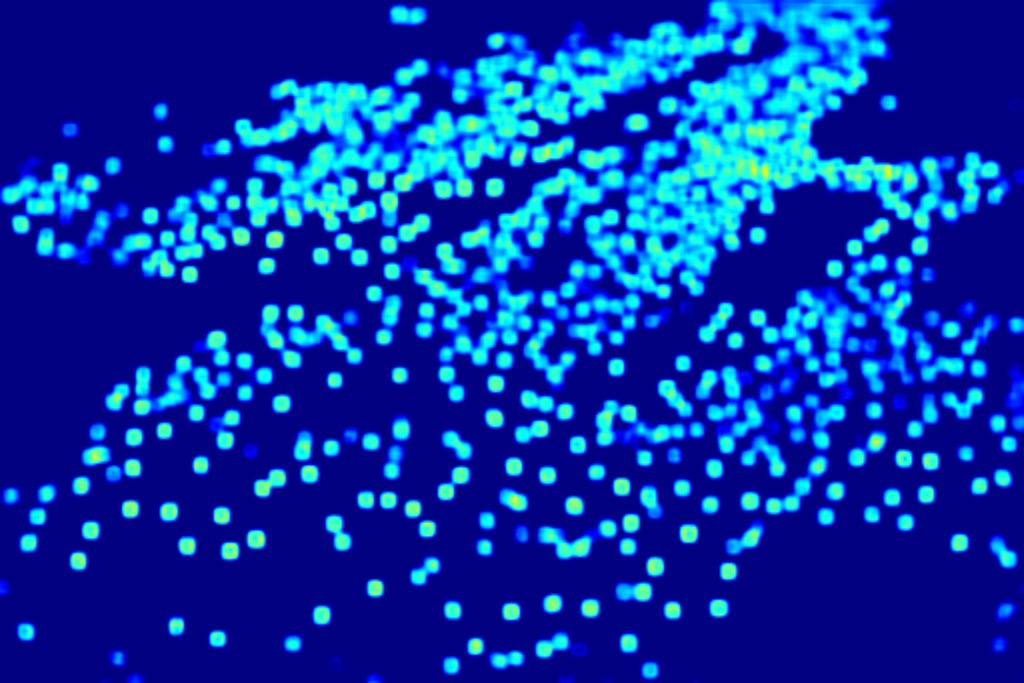}
		\caption{Attention Map}
	\end{subfigure}
	\begin{subfigure}[t]{0.23\linewidth}
		\centering
		\includegraphics[width=1\linewidth]{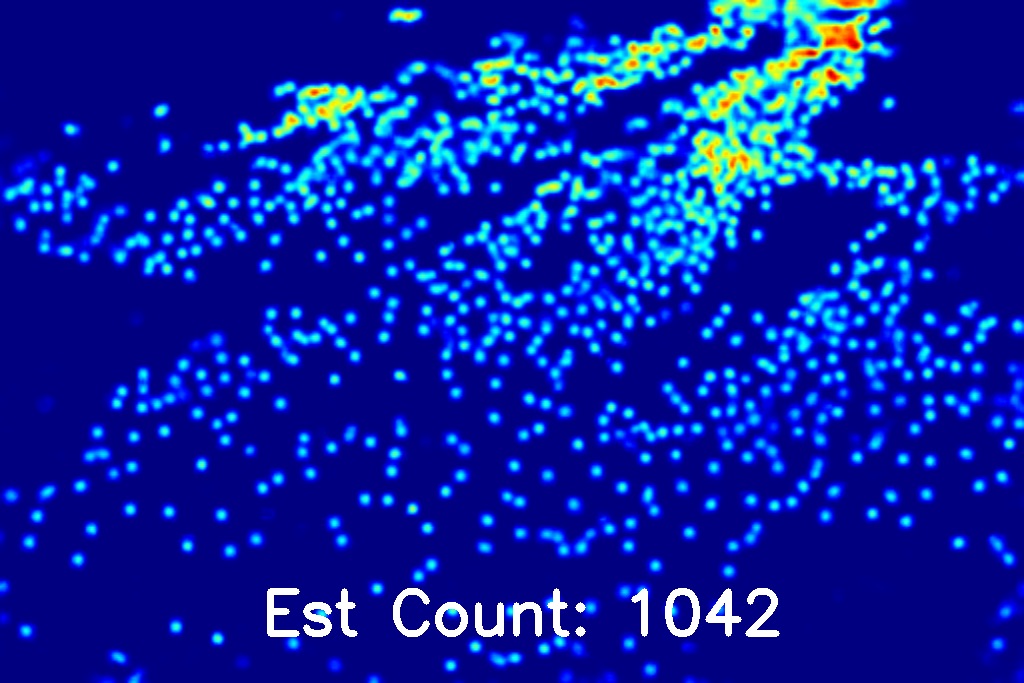}
		\caption{Density Map}
	\end{subfigure}
	\begin{subfigure}[t]{0.23\linewidth}
		\centering
		\includegraphics[width=1\linewidth]{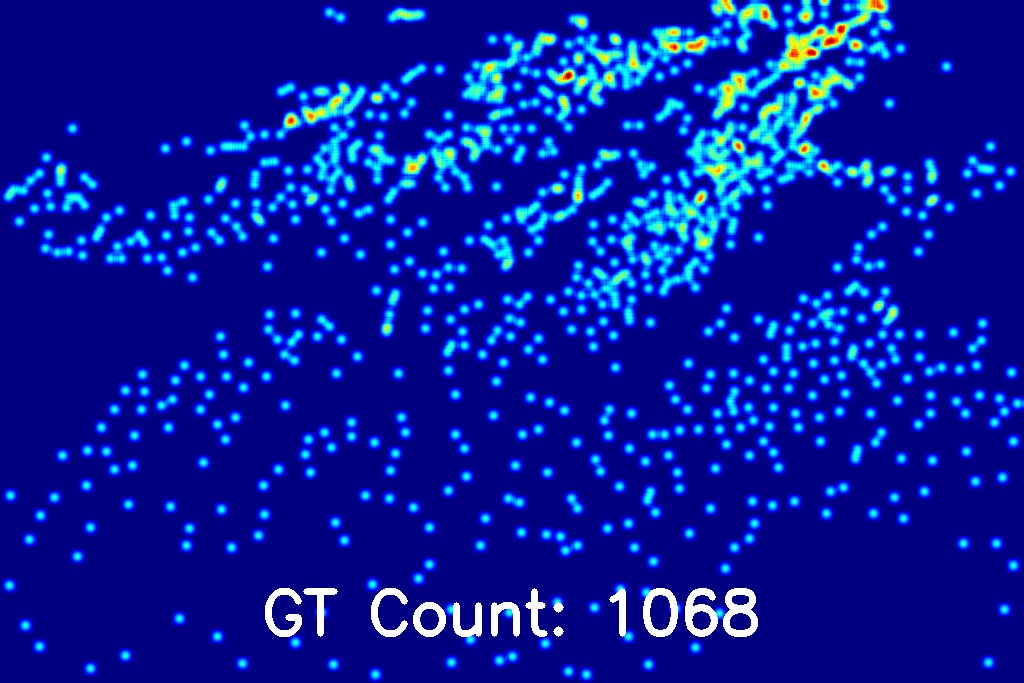}
		\caption{Ground Truth}
	\end{subfigure}
	\caption{Attention maps and density maps of ShanghaiTech dataset generated by SFANet. Row 1 and 2
	are examples from ShanghaiTech part\_B. Row 3 and 4
	are examples from ShanghaiTech part\_A.}\label{figure5}
\end{figure*}

ShanghaiTech crowd counting dataset contains 1198 annotated images with a total amount of 330,165 persons.
This dataset consists of two parts, A and B. Part A contains 482
images with highly congested scenes randomly downloaded from the internet.
Part B contains 716 images with relatively sparse crowd scenes taken from busy streets in Shanghai.
Considering lack of training samples in these datasets, our models are pre-trained on UCF-QNRF,
because it improves performance and speeds up convergence.
We evaluate our method and compare it to other ten
recent works and results are shown in Table\ref{tab1}. It shows that with the novel network 
integrating with attention model, our method achieves the best performance
on both Part\_A and Part\_B datasets among all those approaches. 
Compared to the state-of-the-art method called SaNet, we get $17.9\%$ MAE,
$19.9\%$ MSE improvement for Part\_B. And as far as we know,
our method is the first solution that can break through 8.0 MAE in Part\_B.
More samples of test results can be found in Fig.\ref{figure5}.

\begin{table}[h]
	\centering
\begin{tabular}{|l|c|c|c|c|}
	\hline
	&\multicolumn{2}{|c|}{Part A}&\multicolumn{2}{|c|}{Part B}\\
	\hline
	Method&MAE&MSE&MAE&MSE\\
	\hline
	\hline
	Cross-Scene\cite{zhang2015cross}(2015)&181.8 &277.7&32.0&49.8\\
	\hline
	MCNN\cite{zhang2016single}(2016) &110.2&173.2&26.4&41.3\\
	\hline
	Switch-CNN\cite{sam2017switching}(2017)&90.4&135.0&21.6&33.4\\
	\hline
	CP-CNN\cite{sindagi2017generating}(2017)&73.6&106.4&20.1&30.1\\
	\hline
	TDF-CNN\cite{sam2018top}(2018)&97.5&145.1&20.7&32.8\\
	\hline
	SaCNN\cite{zhang2018crowd}(2018)&86.8&139.2&16.2&25.8\\
	\hline
	ACSCP\cite{shen2018crowd}(2018)&75.7&102.7&17.2&27.4\\
	\hline
	ic-CNN\cite{ranjan2018iterative}(2018)&68.5&116.2&10.7&16.0\\
	\hline
	CSRNet\cite{li2018csrnet}(2018)&68.2&115.0&10.6&16.0\\
	\hline
	SaNet\cite{cao2018scale}(2018)&67.0&104.5&8.4&13.6\\
	\hline
	\textbf{SFANet w/o pretrain}&63.8&105.2&7.6&11.4\\
	\hline
	\textbf{SFANet}&\textbf{59.8}&\textbf{99.3}&\textbf{6.9}&\textbf{10.9}\\
	\hline
\end{tabular}
	\caption{Estimation errors on ShanghaiTech dataset}
\label{tab1}
\end{table}

\subsection{UCF\_CC\_50 dataset}
UCF\_CC\_50 is an extremely dense
crowd dataset introduced by Idrees et al.\cite{idrees2013multi}. It includes 50 images of different
resolutions. The number of annotated
persons per image ranges from 94 to 4543 with an average
number of 1280. For better verification of model accuracy,
5-fold cross-validation is performed following the standard setting in \cite{idrees2013multi}.
Table\ref{tab2} shows the experiment result of MAE and MSE compared with other approaches.

\begin{table}[h]
	\centering
	\begin{tabular}{|l|c|c|}
		\hline
		Method&MAE&MSE\\
		\hline
		\hline
		Cross-Scene\cite{zhang2015cross}&467.0 &498.5\\
		\hline
		MCNN\cite{zhang2016single} &377.6&509.1\\
		\hline
		Switch-CNN\cite{sam2017switching}&318.1&439.2\\
		\hline
		CP-CNN\cite{sindagi2017generating}&295.8&320.9\\
		\hline
		TDF-CNN\cite{sam2018top}&354.7&491.4\\
		\hline
		SaCNN\cite{zhang2018crowd}&314.9&424.8\\
		\hline
		ACSCP\cite{shen2018crowd}&291.0&404.6\\
		\hline
		ic-CNN\cite{ranjan2018iterative}&260.9&365.5\\
		\hline
		CSRNet\cite{li2018csrnet}&266.1&397.5\\
		\hline
		SaNet\cite{cao2018scale}&258.4&334.9\\
		\hline
		\textbf{SFANet}&\textbf{219.6}&\textbf{316.2}\\
		\hline
	\end{tabular}
	\caption{Estimation errors on UCF\_CC\_50 dataset}
	\label{tab2}
\end{table}

\subsection{UCF-QRNF dataset}
UCF-QNRF
dataset\cite{idrees2018composition} is a new and the largest
crowd dataset for evaluating crowd counting and localization methods. 
It contains 1535 dense crowd images which are divided into train and test sets of 
1201 and 334 images respectively. The UCF-QNRF dataset has the most number 
of high-count crowd images and annotations, and a wider variety of scenes containing 
the most diverse set of viewpoints, densities and lighting variations. 
Besides high-density regions, the dataset also contains buildings, vegetation, 
sky and roads as they are present in realistic 
scenarios captured in the wild, which makes this dataset more realistic as well as difficult.
We train model on UCF-QNRF dataset only with VGG16-bn pre-train weights.
Results are shown in Table 4. The SFANet obtains the best
performance with 23.6\% MAE and 8.6\% MSE improvement compared
with the second best approach in \cite{idrees2018composition}.

\begin{table}[h]
	\centering
	\begin{tabular}{|l|c|c|}
		\hline
		Method&MAE&MSE\\
		\hline
		\hline
		Idrees et al. \cite{idrees2013multi}&315.0 &508.0\\
		\hline
		MCNN\cite{zhang2016single} &277.0&426.0\\
		\hline
		CMTL\cite{sindagi2017cnn}&252.0&514.0\\
		\hline
		Switch-CNN\cite{sam2017switching}&228.0&445.0\\
		\hline
		CL-CNN\cite{idrees2018composition}&132.0&191.0\\
		\hline
		\textbf{SFANet}&\textbf{100.8}&\textbf{174.5}\\
		\hline
	\end{tabular}
\caption{Estimation errors on UCF-QRNF dataset}
\label{tab3}
\end{table}

\begin{table}[h]
	\centering
	\begin{tabular}{|l|c|c|}
		\hline
		Method&MAE&MSE\\
		\hline
		\hline
		Cross-Scene\cite{zhang2015cross}&1.60 &3.31\\
		\hline
		MCNN\cite{zhang2016single} &1.07&1.35\\
		\hline
		Switch-CNN\cite{sam2017switching}&1.62&2.10\\
		\hline
		ACSCP\cite{shen2018crowd}&1.04&1.35\\
		\hline
		CSRNet\cite{li2018csrnet}&1.16&1.47\\
		\hline
		SaNet\cite{cao2018scale}&1.02&1.29\\
		\hline
		\textbf{SFANet}&\textbf{0.82}&\textbf{1.07}\\
		\hline
	\end{tabular}
	\caption{Estimation errors on UCSD dataset}
	\label{tab4}
\end{table}

\subsection{UCSD dataset}
The UCSD dataset\cite{chan2008privacy} has 2000 frames taken from a stationary camera.
All videos are 8-bit grayscale, with dimensions $238\times158$ at 10 fps, and the crowd
count in each image varies from 11 to 46. The dataset also provides ROI
region and perspective information. Since the image size is too small to generate 
high-quality density maps, we enlarge each image to $960\times640$ size by bilinear interpolation.
Among the
2000 frames, we use frames 601 through 1400 as training
set and the rest of them as testing set according to\cite{chan2008privacy}. All the frames and density 
maps are masked with ROI. The results
in Table \ref{tab4} indicate that our method can perform 
well not only for
extremely dense crowds but also for sparse
crowds.

\subsection{Ablation Experiments}
Finally, we do ablation experiments to confirm the benefits of attention map path with attention loss.
In experiments, we only use VGG backbone and DMP network noted as VGG-DMP. As the comparison 
result illustrated in Table\ref{tab5}, VGG-DMP also shows its significant 
performance improvement and surpass the other state-of-the-art methods 
even without attention map path. The experiment results
also confirm a notable effect of attention map path as expected.

\begin{table}[h]
	\centering
	\begin{tabular}{|l|c|c|c|c|}
		\hline
		&\multicolumn{2}{|c|}{Part A}&\multicolumn{2}{|c|}{Part B}\\
		\hline
		Method&MAE&MSE&MAE&MSE\\
		\hline
		\hline
		CSRNet\cite{li2018csrnet}&68.2&115.0&10.6&16.0\\
		\hline
		SaNet\cite{cao2018scale}&67.0&107.5&8.4&13.6\\
		\hline
		\textbf{VGG-DMP}&62.7&107.0&7.8&12.7\\
		\hline
		\textbf{SFANet}&\textbf{59.8}&\textbf{99.3}&\textbf{6.9}&\textbf{10.9}\\
		\hline
	\end{tabular}
	\caption{Ablation experiments on the ShanghaiTech dataset}
	\label{tab5}
\end{table}

\section{Conclusion}
We proposed a novel end-to-end model, named SFANet,
based on dual path multi-scale fusion with attention mechanism for
crowd counting. Multi-scale feature maps are extracted by VGG16-bn backbone,
and fused by DMP and AMP. Attention map loss is well designed to emphasize
the head regions among noisy background. By taking these two advantages,
SFANet shows powerful ability to locate the head regions and regress the head count.
Experiments indicate that SFANet
achieves the best performance and higher robustness
for crowd counting compared with other
state-of-the-art approaches.

{\small
\bibliographystyle{ieee}
\bibliography{egbib}
}

\end{document}